\documentclass{llncs}

\usepackage{amsmath, amsfonts, amssymb}
\usepackage{graphicx}
\usepackage{url}
\usepackage{blindtext}
\usepackage{enumitem}
\usepackage{graphicx}
\usepackage{amssymb}
\usepackage{amsmath}
\usepackage{algpseudocode}
\usepackage{caption}
\usepackage{subfig}
\usepackage[sort,nocompress]{cite}
\usepackage{hhline}
\usepackage{authblk}
\usepackage{comment}
\usepackage[export]{adjustbox}
\usepackage{hyperref}
\usepackage{tikz}
\usetikzlibrary{decorations.pathreplacing}
\usetikzlibrary{calc}
\usepackage{scalerel}
\usepackage{amsmath}
\usepackage{graphicx}
\usepackage{caption}
\usepackage{multirow}
\usepackage{newtxtext,newtxmath}
\usepackage{makecell}

\usetikzlibrary{arrows,positioning,calc,topaths, fit, backgrounds}
  \tikzset{
    annotated cuboid/.pic={
      \tikzset{%
        every edge quotes/.append style={midway, auto},
        /cuboid/.cd,
        #1
      }
      \draw [\cubeline,every edge/.append style={pic actions,\cubeback, opacity=.5}, pic actions]
      (0,0,0) coordinate (o-\cubelabel) -- ++(-\cubescale*\cubex,0,0) coordinate (a-\cubelabel) -- ++(0,-\cubescale*\cubey,0) coordinate (b-\cubelabel) edge coordinate [pos=1] (g-\cubelabel) ++(0,0,-\cubescale*\cubez)  -- ++(\cubescale*\cubex,0,0) coordinate (c-\cubelabel) -- cycle
      (o-\cubelabel) -- ++(0,0,-\cubescale*\cubez) coordinate (d-\cubelabel) -- ++(0,-\cubescale*\cubey,0) coordinate (e-\cubelabel) edge (g-\cubelabel) -- (c-\cubelabel) -- cycle
      (o-\cubelabel) -- (a-\cubelabel) -- ++(0,0,-\cubescale*\cubez) coordinate (f-\cubelabel) edge (g-\cubelabel) -- (d-\cubelabel) -- cycle;

    },
    /cuboid/.search also={/tikz},
    /cuboid/.cd,
    width/.store in=\cubex,
    height/.store in=\cubey,
    depth/.store in=\cubez,
    units/.store in=\cubeunits,
    scale/.store in=\cubescale,
    label/.store in=\cubelabel,
    line/.store in=\cubeline,
    backline/.store in=\cubeback,
    width=10,
    height=10,
    depth=10,
    units=cm,
    scale=.1,   
    line=draw,
    backline=densely dashed
  }
  
\newcommand{\cuboid}[2]{
    \begin{tikzpicture}
     \pic [#1] at (0,0) {annotated cuboid={#2}};
    \end{tikzpicture}
}

\definecolor{conv_color}{rgb}{0.67, 0.94, 0.82}
\definecolor{bn_relu_color}{rgb}{1.0, 0.85, 0.73}
\definecolor{pooling_color}{rgb}{0.9, 0.89, 0.89}
\definecolor{dense_block_color}{rgb}{0.67, 0.9, 0.93}
\definecolor{transit_color}{rgb}{0.94, 0.92, 0.84}
\definecolor{deconv_color}{rgb}{0.6, 1.0, 0.6}
\definecolor{classify_color}{rgb}{0.8, 0.8, 1.0}
\definecolor{softmax_color}{rgb}{0.6, 0.81, 0.93}
\definecolor{auxiliary_color}{rgb}{0.3, 0.3, 0.5}
\definecolor{copy_color}{rgb}{0.0, 0.75, 1.0}
\definecolor{concat_color}{rgb}{0.7, 0.5, 0.5}
\definecolor{defination_color}{rgb}{0.94, 0.97, 1.0}
\newcommand\widthpatch{1.5}
\newcommand\widthconv{2}
\newcommand\widthpooling{1.5}
\newcommand\widthdensesecond{4}
\newcommand\widthtransitsecond{2}
\newcommand\widthdensethird{4}
\newcommand\widthtransitthird{2}
\newcommand\widthdensefourth{4}
\newcommand\widthtransitfourth{2}
\newcommand\widthdensefifth{4}
\newcommand\widthdeconvfirst{2}
\newcommand\widthdeconvsecond{1}
\newcommand\widthdeconvthird{1}
\newcommand\widthdeconvfourth{1}
\newcommand\widthdeconvfifth{1}

\newcommand\convlocationx{2}
\newcommand\marginconv{0.33}
\tikzstyle{cubecontainer}=[outer sep = 0pt, inner sep= 0pt,]
\tikzstyle{connectarrow}=[-{Triangle[angle=60:0pt 2]},
                          line width= 8pt, shorten >=2mm,shorten <=2mm, draw=green!60]

\begin{document}

\mainmatter  

\title{3D Densely Convolutional Networks for Volumetric Segmentation}
\author{Toan Duc Bui, Jitae Shin, and Taesup Moon}
\institute{School of Electronic and Electrical Engineering, Sungkyunkwan University, Republic of Korea}

\maketitle

\begin{abstract}
In the isointense stage, the accurate volumetric image segmentation is a challenging task due to the low contrast between tissues. In this paper, we propose a novel very deep network architecture based on densely convolutional network for volumetric brain segmentation. The proposed network architecture provides a dense connection between layers that aims to improve the information flow in the network. By concatenating features map of fine and coarse dense blocks, it allows capturing multi-scale contextual information. Experimental results demonstrate significant advantages of the proposed method over existing methods, in terms of both segmentation accuracy and parameter efficiency in  MICCAI grand challenge on 6-month infant brain MRI segmentation. 
\end{abstract}

\section{Introduction}

Volumetric brain image segmentation aims to separate the brain tissues into non-overlapping regions such as white matter (WM), gray mater (GM), cerebrospinal fluid (CSF) and background (BG) regions. The accurate volumetric image segmentation is a prerequisite for quantifying the structural volumes. However, the low contrast between tissues often cause tissue to be misclassified, which can hinder accurate segmentation. Hence, the accuracy of automatic brain segmentation is still an active area of research. 

Recently, deep convolutional neural networks \cite{lecun1998gradient} have achieved a great success in medical image segmentation \cite{zhang2015deep, ronneberger2015u, moeskops2016automatic}. For example, {\c{C}}i{\c{c}}ek et al. \cite{cciccek20163d} proposed a 3D U-net that contracting skip layers and learned up-sampling part to produce a full-resolution segmentation. Chen et al. \cite{chen2017voxresnet} proposed a voxelwise residual network for brain segmentation by pass signal from one layer to the next via identity connection. However, the connection is a short path from early layers to later layer. To address it, Huang \cite{huang2017densely} introduces a DenseNet that provides a direct connections from any layer to all subsequent layers to ensure maximum information flow between layers. It shows a consistent improvement in accuracy with increasing depth network. Yu \cite{yu2017automatic} extended DenseNet to volumetric cardiac segmentation. It uses two dense blocks and follows by pooling layers to reduce feature maps resolution, then restores the resolution by stacks of learned deconvolution layers. These stack deconvolution layers often generate a larger learned parameters that take a lot of memory for the training process. It also may not be able to capture multi-scale contextual information, results in a poor performance. 

In this paper, we propose a novel very deep network architecture based on densely convolution network for volumetric brain segmentation. First, we combine local predictions and global predictions by concatenating features map of fine and coarse dense blocks that allow capturing multi-scale contextual information. In the traditional DenseNet architecture \cite{huang2017densely}, the pooling layer often uses to reduce feature resolution and to increase the abstract feature representations; however, it may lose the spatial information. To preserve the spatial information, we replace the pooling layer with a convolution layer of stride 2. It only increases a small number of learned parameters, but significantly improves the performance.  Second, we use a model of bottleneck with compression (BC) to reduce the number of feature maps in each dense block to reduce the number of learned parameters results in computational efficiency than existing methods \cite{cciccek20163d,yu2017automatic}. Experimental results demonstrate significant advantages of the proposed method over existing methods, in terms of both segmentation accuracy and parameter efficiency in  MICCAI grand challenge on 6-month infant brain MRI segmentation\footnote{http://iseg2017.web.unc.edu/}. Our implementation and network architectures are publicly available at the website\footnote{https://github.com/tbuikr/3D\_DenseSeg}

\section{Methods}

In this section, we first briefly review the key concept of DenseNet \cite{huang2017densely} to deal with the degradation problem in the classification task. Then, we propose a novel network architecture that extends the DenseNet to volumetric segmentation.

\subsection{DenseNet: Densely Connected Convolutional Network}
Let $x_\ell$ be the output of $\ell^{th}$ layer. In the traditional feed-forward networks, the  output $x_\ell$ is computed by
\begin{equation}
    x_\ell=\mathit{H_{\ell}}(x_{\ell-1})
\end{equation}
where $\mathit{H}$ is a non-linear transformation of $\ell^{th}$ layer. In particular, the performance of the deep network architecture gets saturated with the network depth increasing due to vanishing/exploding gradient \cite{he2016deep}. To address the degradation problem, Huang \cite{huang2017densely} introduced a DenseNet architecture that provides a direct connection from any layer to subsequent layers by concatenation the feature maps of all preceding layers as follows:

\begin{equation}
    x_\ell=\mathit{H_\ell}([x_{0}, x_{1}, \cdots,  x_{\ell-1}])
\end{equation}
where $[.]$ denotes the concatenation operation that concates the feature maps of all subsequent layers. 

By using dense connections, the DenseNet architecture allows better information and gradient flow during training. If each function $\mathit{H_\ell}(.)$ produces $k$ feature maps as output, then the number of input feature map at the layer $\ell^{th}$ will be $k_0+(\ell -1)\times k$, where $k_0$ is a number of feature map at the first layer.  The hyper-parameter $k$ refers as growth rate. To reduce the the number of the input feature maps, a $1 \times 1$ convolution layer is introduced as bottleneck layer before each $3 \times 3$ convolution. To further improve model compactness, a transition layer that includes a batch normalization layer (BN) \cite{ioffe2015batch}, a ReLU \cite{glorot2011deep} and a $1 \times 1$ convolutional layer followed by $2 \times 2$ pooling layer \cite{lecun1998gradient}, is used to reduce the feature maps resolution. With $m$ input feature maps, the transition layer generates $m\times\theta$ output feature maps, where $0 \le \theta \le 1$. The network architecture is referred as DenseNet-BC when the bottleneck and transition layers with $\theta \le 1$ are used.

\subsection{3D-DenseSeg: A 3D Densely Convolution Networks for Volumetric Segmentation}
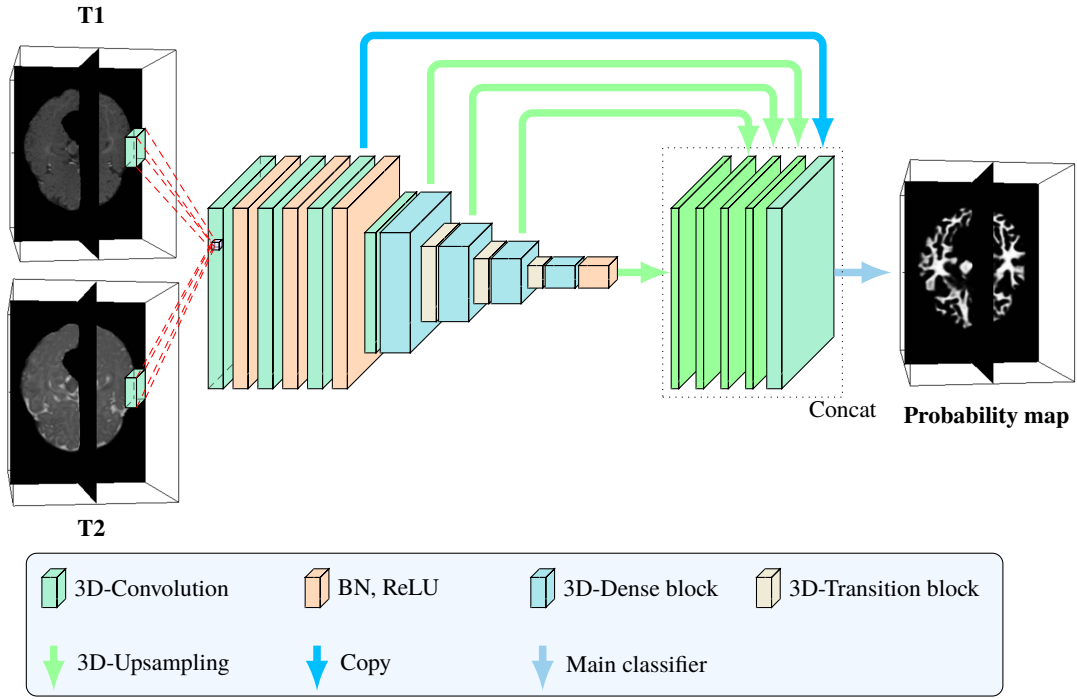
\begin{figure*} [h]
\centering
\begin{tikzpicture}
\begin{scope}[local bounding box=scope1]

\node [inner sep=0pt] at (0,1.6) (inputT1) {\includegraphics[width=.2\textwidth]{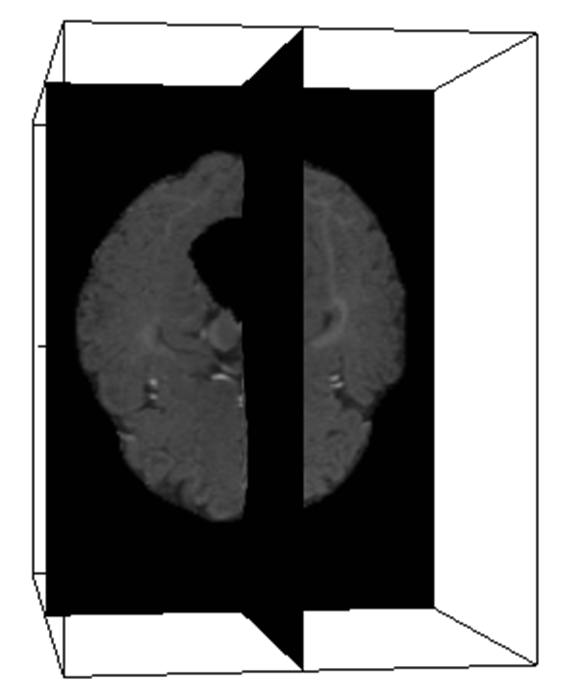}};
\node[above,above=0.1cm](input_T1_title) at (inputT1.north) {\textbf{T1}};
\pic [fill=conv_color, text=green!50!black, draw=black] at (0.6,1.8) {annotated cuboid={label=B, width=\widthpatch, height=4, depth=3, units=m}};

\node [inner sep=0pt] at (0,-1.6) (inputT2) {\includegraphics[width=.2\textwidth]{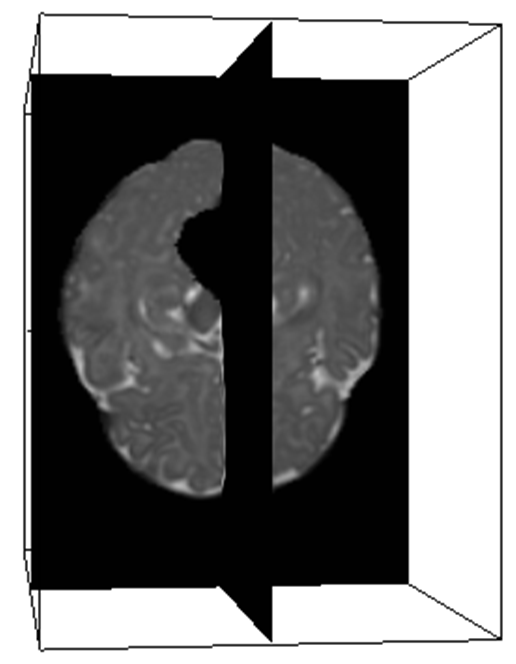}};
\node[above,below=0.0cm](input_T2_title) at (inputT2.south) {\textbf{T2}};
\pic [fill=conv_color, text=green!50!black, draw=black] at (0.6,-1.4) {annotated cuboid={label=C, width=\widthpatch, height=4, depth=3, units=m}};

\node[cubecontainer] (conv1a) at (\convlocationx,0.0) {\cuboid{fill=conv_color, text=green!50!black, draw=black} {label=conv1a, width=\widthconv, height=24, depth=18, units=mm}};

\pic [fill=blue!10, text=green!50!black, draw=black] at (1.7,0.4) {annotated cuboid={label=D, width=1, height=1, depth=1, units=m}};

\node[cubecontainer] (BN_ReLU) at (\convlocationx+\marginconv,0.0) {\cuboid{fill=bn_relu_color, text=green!50!black, draw=black} {label=BN_ReLU, width=\widthconv, height=24, depth=18, units=mm,backline=bn_relu_color}};

\node[cubecontainer] (conv1b) at (\convlocationx+ 2*\marginconv,0.0) {\cuboid{fill=conv_color, text=green!50!black, draw=black} {label=conv1b, width=\widthconv, height=24, depth=18, units=mm,backline=conv_color}};

\node[cubecontainer] (BN_ReLU) at (\convlocationx+3*\marginconv,0.0) {\cuboid{fill=bn_relu_color, text=green!50!black, draw=black} {label=BN_ReLU, width=\widthconv, height=24, depth=18, units=mm,backline=bn_relu_color}};

\node[cubecontainer] (conv1c) at (\convlocationx+4*\marginconv,0.0) {\cuboid{fill=conv_color, text=green!50!black, draw=black} {label=conv1c, width=\widthconv, height=24, depth=18, units=mm, backline=conv_color}};

\node[cubecontainer] (BN_ReLU) at (\convlocationx+5*\marginconv,0.0) {\cuboid{fill=bn_relu_color, text=green!50!black, draw=black} {label=BN_ReLU, width=\widthconv, height=24, depth=18, units=mm,backline=bn_relu_color}};

\node[cubecontainer] (pooling1) at (\convlocationx+6*\marginconv,0.0) {\cuboid{fill=conv_color, text=green!50!black, draw=black}{label=pooling1, width=\widthpooling, height=16, depth=14, units=m, backline=conv_color}};

\node[cubecontainer] (dense_block2) at (\convlocationx+7*\marginconv,0.0) {\cuboid{fill=dense_block_color, text=green!50!black, draw=black}{label=dense_block2, width=\widthdensesecond, height=16, depth=14, units=m, backline=dense_block_color}};

\node[cubecontainer] (transit2) at (\convlocationx+8*\marginconv,0.0) {\cuboid{fill=transit_color, text=green!50!black, draw=black}{label=transit2, width=\widthtransitsecond, height=10, depth=8, units=m, backline=transit_color}};

\node[cubecontainer] (dense_block3) at (\convlocationx+9*\marginconv,0.0) {\cuboid{fill=dense_block_color, text=green!50!black, draw=black}{label=dense_block3, width=\widthdensethird, height=10, depth=8, units=m,backline=dense_block_color}};

\node[cubecontainer] (transit3) at (\convlocationx+10*\marginconv,0.0) {\cuboid{fill=transit_color, text=green!50!black, draw=black}{label=transit3, width=\widthtransitthird, height=6, depth=6, units=m,backline=transit_color}};

\node[cubecontainer] (dense_block4) at (\convlocationx+11*\marginconv,0.0) {\cuboid{fill=dense_block_color, text=green!50!black, draw=black}{label=dense_block4, width=\widthdensefourth, height=6, depth=6, units=m,backline=dense_block_color}};

\node[cubecontainer] (transit4) at (\convlocationx+12*\marginconv,0.0) {\cuboid{fill=transit_color, text=green!50!black, draw=black}{label=transit4, width=\widthtransitfourth, height=3, depth=3, units=m, backline=transit_color}};

\node[cubecontainer] (dense_block5) at (\convlocationx+13*\marginconv,0.0) {\cuboid{fill=dense_block_color, text=green!50!black, draw=black}{label=dense_block5, width=\widthdensefifth, height=3, depth=3, units=m,backline=dense_block_color}};

\node[cubecontainer] (BN_ReLU_last) at (\convlocationx+14.35*\marginconv,0.0) {\cuboid{fill=bn_relu_color, text=green!50!black, draw=black} {label=BN_ReLU_last, width=\widthdensefifth, height=3, depth=3, units=m,backline=bn_relu_color}};


\node[cubecontainer] (deconv5) at (\convlocationx+18.5*\marginconv,0.0) {\cuboid{fill=deconv_color, text=green!50!black, draw=black}{label=deconv5, width=\widthdeconvfifth, height=24, depth=18, units=mm,backline=deconv_color}};

\node[cubecontainer] (deconv4) at (\convlocationx+19.5*\marginconv,0) {\cuboid{fill=deconv_color, text=green!50!black, draw=black}{label=deconv4, width=\widthdeconvfourth, height=24, depth=18, units=mm,backline=deconv_color}};

\node[cubecontainer] (deconv3) at (\convlocationx+20.5*\marginconv,0.0) {\cuboid{fill=deconv_color, text=green!50!black, draw=black}{label=deconv3, width=\widthdeconvthird, height=24, depth=18, units=mm,backline=deconv_color}};

\node[cubecontainer] (deconv2) at (\convlocationx+21.5*\marginconv,0) {\cuboid{fill=deconv_color, text=green!50!black, draw=black}{label=deconv2, width=\widthdeconvsecond, height=24, depth=18, units=mm,backline=deconv_color}};

\node[cubecontainer] (conv_copy) at (\convlocationx+22.5*\marginconv,0.0) {\cuboid{fill=conv_color, text=green!50!black, draw=black}{label=deconv1, width=\widthdeconvfirst, height=24, depth=18, units=mm,backline=conv_color}};

\node [inner sep=1pt,align=center] at (\convlocationx+30*\marginconv,0) (prob_map) {\includegraphics[width=.2\textwidth]{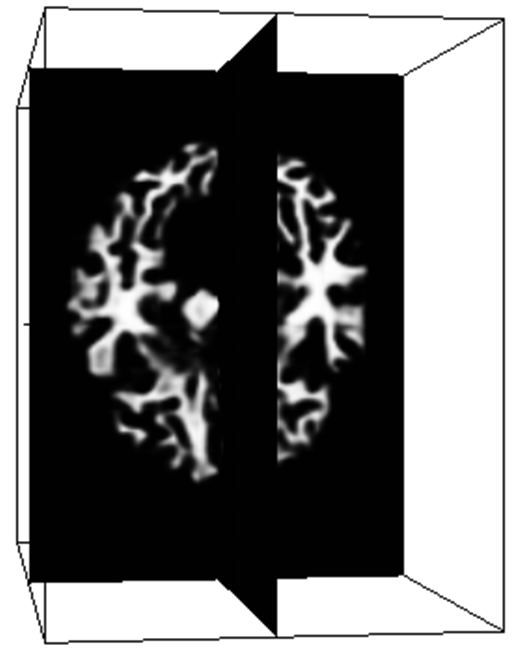}};
\node[above,below=0.1cm](prob_title) at (prob_map.south) {\textbf{Probability map}};

\draw [ densely dashed, draw=red] (e-B) -- (g-D);
\draw [ densely dashed, draw=red] (d-B) -- (f-D);
\draw [ densely dashed, draw=red] (c-B) -- (b-D);
\draw [ densely dashed, draw=red] (o-B) -- (a-D);

\draw [ densely dashed, draw=red] (e-C) -- (g-D);
\draw [ densely dashed, draw=red] (d-C) -- (f-D);
\draw [ densely dashed, draw=red] (c-C) -- (b-D);
\draw [ densely dashed, draw=red] (o-C) -- (a-D);

 
\draw[line width=1mm, copy_color, -latex, shorten >=1mm, shorten <=1mm, rounded corners] ([yshift=-0mm, xshift=3mm] conv1c.north) --++(90:16mm)-|([yshift=-0mm, xshift=3mm] conv_copy.north);

\draw[line width=1mm, deconv_color, -latex, shorten >=1mm, shorten <=1mm, rounded corners] ([yshift=-0mm, xshift=2mm]dense_block2.north) --++(90:17mm)-|([yshift=-0mm, xshift=3mm] deconv2.north);

\draw[line width=1mm, deconv_color, -latex, shorten >=1mm, shorten <=1mm, rounded corners] ([yshift=-0mm, xshift=1mm]dense_block3.north) --++(90:18mm)-|([yshift=-0mm, xshift=3mm] deconv3.north);

\draw[line width=1mm, deconv_color, -latex, shorten >=0mm, shorten <=1mm, rounded corners] ([yshift=-0mm, xshift=1mm]dense_block4.north) --++(90:17mm)-|([yshift=-0mm, xshift=3mm] deconv4.north);

\draw[line width=1mm, deconv_color, -latex, shorten >=0mm, shorten <=0mm] (BN_ReLU_last.east) -- (deconv5.west);

\draw[line width=1mm, softmax_color, -latex, shorten >=0mm, shorten <=0mm] (conv_copy.east) -- (prob_map.west);


\node[draw,dotted,fit=(deconv5) (deconv4) (deconv3) (deconv2) (conv_copy)] (concate_node){};
\node[below=1.6](concate_node) at (concate_node.east) {Concat};

\end{scope}

\begin{scope}
\node[cubecontainer] (conv_def) at (-0.5,-4.2) {\cuboid{fill=conv_color, text=green!50!black, draw=black} {label=conv_def, width=2, height=4, depth=3, units=m,backline=conv_color}};
\node (conv_def_text) [right= 0.0 of conv_def] {3D-Convolution}; 

\node[cubecontainer] (bn_def) at (3.,-4.2) {\cuboid{fill=bn_relu_color, text=green!50!black, draw=black} {label=bn_def, width=2, height=4, depth=3, units=m,backline=bn_relu_color}};
\node (bn_def_text) [right= 0.0 of bn_def] {BN, ReLU}; 

\node[cubecontainer] (dense_def) at (6,-4.2) {\cuboid{fill=dense_block_color, text=green!50!black, draw=black} {label=conv_def, width=2, height=4, depth=3, units=m,backline=dense_block_color}};
\node (conv_def_text) [right= 0.0 of dense_def] {3D-Dense block};

\node[cubecontainer] (transit_def) at (9,-4.2) {\cuboid{fill=transit_color, text=green!50!black, draw=black} {label=conv_def, width=2, height=4, depth=3, units=m,backline=transit_color}};
\node (transit_def_text) [right= 0.0 of transit_def] {3D-Transition block}; 

\coordinate [below=0.5cm of conv_def] (deconv_coordinate);
\draw[line width=1.0mm, deconv_color, -latex, shorten >=0mm, shorten <=0mm] ([yshift=-0mm, xshift=-0mm]deconv_coordinate.south west)--++(-90:6mm);
\node (deconv_def_text) [right= 0.2 of deconv_coordinate,yshift=-2.5mm] {3D-Upsampling}; 

\coordinate [below=0.5cm of bn_def] (auxility_classify_def);
\draw[line width=1.0mm, copy_color, -latex, shorten >=0mm, shorten <=0mm] ([yshift=-0mm, xshift=-0mm]auxility_classify_def.south west)--++(-90:6mm);
\node (auxility_classify_def) [right= 0.2 of auxility_classify_def,yshift=-2.5mm] {Copy}; 

\coordinate [below=0.5cm of dense_def] (classify_def);
\draw[line width=1.0mm, softmax_color, -latex, shorten >=0mm, shorten <=0mm] ([yshift=-0mm, xshift=-0mm]classify_def.south west)--++(-90:6mm);
\node (classify_def_text) [right= 0.2 of classify_def,yshift=-2.5mm] {Main classifier};

 \begin{scope}[on background layer]
\node [fit= (conv_def) (classify_def_text) (dense_def) (transit_def_text), fill=defination_color, draw, rounded corners, inner sep=.2cm, align=center] (node_fit_define){};   
 \end{scope}
 \end{scope}
\end{tikzpicture}

\caption{3D-DenseSeg network architecture for volumetric segmentation} \label{fig:proposed_network_architecture}
\end{figure*}

Fig. \ref{fig:proposed_network_architecture} illustrates the proposed network architecture for volumetric segmentation. It consists of 47 layers with 1.55 million learned parameters. The network includes two paths: down-sampling and up-sampling. The down-sampling aims to reduces the feature maps resolution, and to increase the receptive field. It is performed by four dense blocks, in which each dense block consists of four BN-ReLU-Conv($1\times1\times1$)-BN-ReLU-Conv($3\times3\times3$) with growth rate $k=16$. We use a dropout layer \cite{srivastava2014dropout} with the dropout rate of $0.2$ after each Conv($3\times3\times3$) layer in the dense block to against the over-fitting problem. Between two contiguous dense blocks,  a transition block includes Conv($1\times1\times1$) with $\theta=0.5$ followed by a convolution layer of stride 2 is used to reduce the feature maps resolution, while preserving the spatial information. The transition block serves as a deep supervision to handle with the limited dataset, but less complicated and ignoring the tuning weight balance between the auxiliary and main losses. Before entering the first dense block, we extract feature by using three convolution layers that generate $k_0=32$ output feature maps.

In the up-sampling path, the 3D-Upsampling operators are used to recover the input resolution. In particular, the shallower layers contains the local feature, while the deeper layer contains the global feature \cite{long2015fully}. To make a better prediction, we perform up-sampling after each dense block and combine these up-sampling feature maps. The concatenation from the different level of up-sampling feature maps allows capturing multiple contextual information. A classifier consisting of a  Conv($1\times 1\times 1$) is used to classify the concatenation feature maps into target classes (i.e four classes for the brain). Finally, the brain probability maps can be obtained using softmax classification. 

\section{Experiments}
\subsection{Dataset and Training} We used the public 6-month infant brain MRI segmentation challenge (iSeg) dataset\footnote{http://iseg2017.web.unc.edu/} to evaluate the proposed method. It consists of 10 training samples and 13 testing samples. Each sample includes a T1 image, a T2 image that were performed pre-processing using in-house tools. Automatic segmentations will be compared with the manual segmentation, by using various measurements, such as Dice Coefficient (DC), Modified Hausdorff Distance (MHD) and Average Surface Distance (ASD).

 We normalize the T1 and T2 images to zero mean and unit variance before entering them into our network. Our network is trained by Adam method \cite{kingma2014adam} with a mini-batch size of $4$. The weights is initialized as in He et. al \cite{he2015delving}. The learning rate was initially set to 0.0002, and drop learning rate by a factor of $\gamma=0.1$ every 50000 iterations. We used weight decay of 0.0005 and a momentum of 0.97. Due to the limited GPU memory, we randomly cropped sub-volume samples with size of $64\times 64 \times 64$ for the input of the network. We used voting strategy to generate the final segmentation results from the predictions of the overlapped sub-volumes. 
\subsection{Performance}
\subsubsection{Evaluation proposed method} 
\begin{figure*}[ht]
   \centering
    \includegraphics[width=1\textwidth]{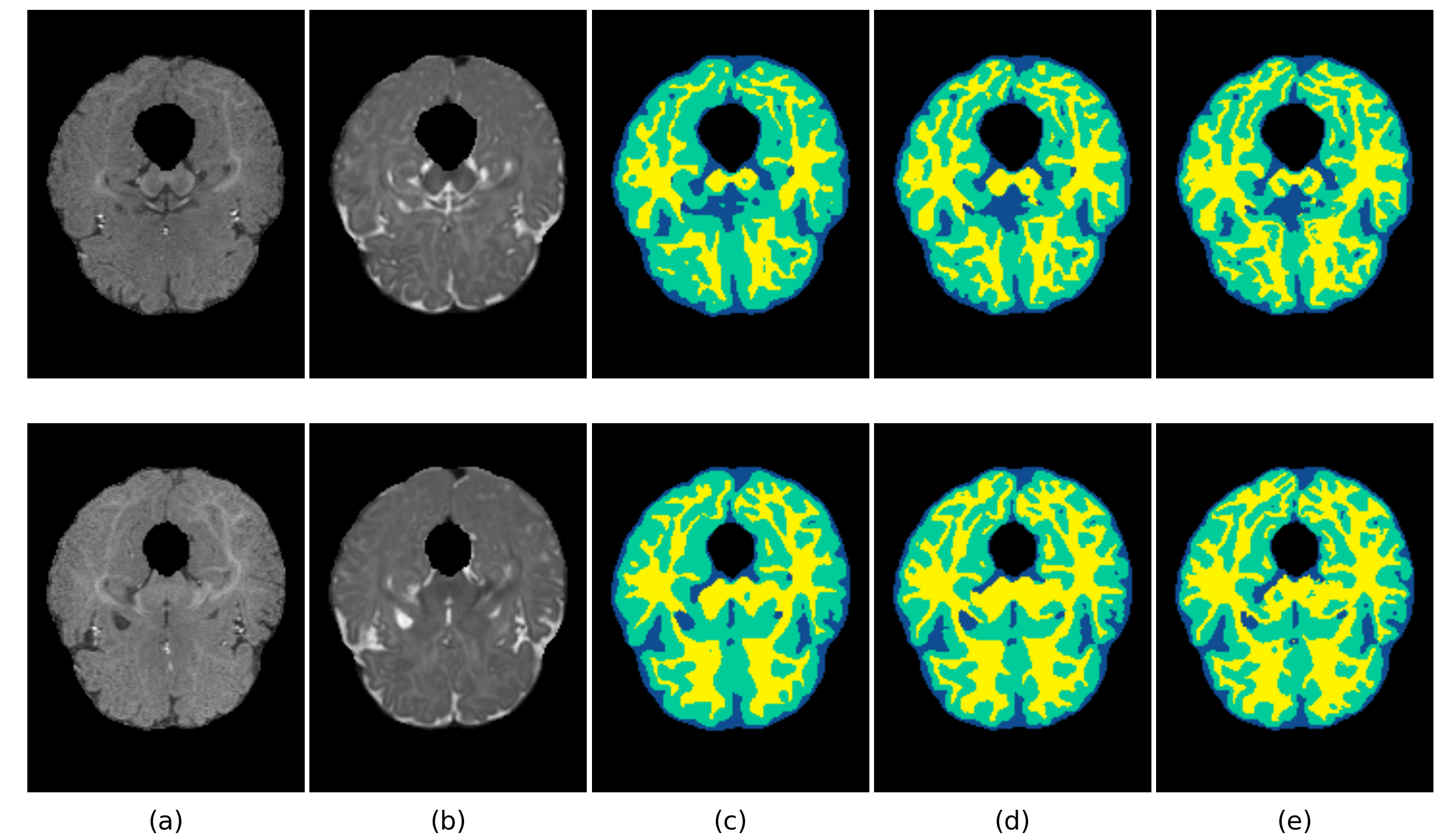}%
\caption{Segmentation result on different slice (a) T1 image, (b) T2 image, (c) DenseVoxNet, (d) our result, (e) manual segmentation}
   \label{fig:validation}%
\end{figure*}

To evaluate the performance of the proposed method, we perform cross-validation on iSeg dataset. Fig. \ref{fig:validation} shows the validation results of the proposed method for the ninth subject on different slices. This demonstrates the robustness of the proposed network architecture for accurate segmentation. 
\subsubsection{Comparison with the existing methods} In order to evaluate the proposed method quantitatively, we compare the proposed method with the state-of-the-art deep learning based methods \cite{cciccek20163d},  \cite{yu2017automatic} on validation set in term of the number of depth, learned parameters and segmentation accuracy, as shown in Table ~\ref{table_compare}. The 3D-Unet \cite{cciccek20163d} architecture consisting of 18 layers with 19 million learned parameters achieves $91.58$\% accuracy. The state-of-the-art DenseVoxNet \cite{yu2017automatic} based on DenseNet architecture with stacked deconvolutions has 32 layers with 4.34 million learned parameters achieved 89.23\% accuracy. The proposed network architecture provides substantially deeper networks (47 layers), while it only has 1.55 million learned parameters due to using BC model. It achieved $92.50\%$ accuracy with less learned parameters in comparison with existing methods. Table ~\ref{table_test} show the results on the test set of the iSeg dataset\footnote{http://iseg2017.web.unc.edu/results/}. It is observed that the proposed network architecture achieved state-of-the-art performance in the challenge.

\begin{table}[!t]
\renewcommand{\arraystretch}{1.3}
\setcellgapes{3pt}\makegapedcells
\caption{A comparison between proposed method with state-of-the-art method in term of parameter and accuracy on the validation}
\label{table_compare}
\centering
\begin{tabular}{l|c|c|c|c|c|c}
\hline
\multicolumn{1}{c|} {\multirow{2}{*}{ \bfseries Method}}  & \multicolumn{1}{c|}{\multirow{2}{*}{ \bfseries Depth}} &\multicolumn{1}{c|}{\multirow{2}{*}{ \bfseries Params}} &\multicolumn{3}{c|}{\bfseries DSC} 
&\multicolumn{1}{c}{\multirow{2}{*}{ \bfseries Average DSC}} \\
\cline{4-6}
&&&WM & GM & CSF \\
\hline
3D-Unet (2015)& 18 & 19M&89.57&90.73&94.44&91.58 \\ 
DenseVoxNet (2017)& 32 & 4.34M &85.46&88.51& 93.71&89.23 \\  
3D-DenseNet (Ours) & 47 & 1.55M &\textbf{91.25} & \textbf{91.57} & \textbf{94.69}& \textbf{92.50}  \\  
\hline
\end{tabular}    
\end{table}
\begin{table*}[!t]
\renewcommand{\arraystretch}{1.3}
\setcellgapes{3pt}\makegapedcells
\caption{Results of iSeg-2017 challenge of different methods (DC:\%, MHD: mm,
ASD: mm. only top 5 teams are shown here).}
\label{table_test}
\centering
\begin{tabular}{l|c|c|c|c|c|c|c|c|c}
\hline
\multicolumn{1}{c|} {\multirow{2}{*}{ \bfseries Method}}  & \multicolumn{3}{c|}{\bfseries WM} &\multicolumn{3}{c|}{\bfseries GM} &\multicolumn{3}{c}{\bfseries CSF} \\
\cline{2-10}
&DSC & MHD & ASD &DSC & MHD & ASD &DSC & MHD & ASD\\
\hline
3D-DenseSeg (MSL\_SKKU-Ours) &\textbf{90.1}&    \textbf{6.444}    &0.391&        \textbf{91.9}&    5.980&    \textbf{0.330}&    \textbf{    95.8}&    9.072&    \textbf{0.116}\\  
LIVIA &89.7    &6.975&    \textbf{0.376}&        91.9    &6.415&    0.338&        95.7&    \textbf{9.029}&    0.138  \\  
Bern\_IPMI & 89.6    &6.782&    0.398&    91.6&    6.455&    0.341&    95.4&    9.616&    0.127  \\  
LRDE&86.1&    6.607&    0.523&        88.7&    \textbf{5.852}&    0.458    &    92.8    &9.875&    0.201\\

nic\_vicorob&88.5    &7.154    &0.430    &    91.0&    7.647    &0.367        &95.1    &9.178    &0.137\\
\hline
\end{tabular}    
\end{table*}

\section{Discussion and Conclusion}
We have proposed a novel 3D dense network architecture to addresse challenges in volumetric medical segmentation, especially infant brain segmentation. By concatenation information from coarse to fine layers, the proposed network architecture allows to capture multiple contextual information. The proposed network is much deeper than the existing method, and hence can capture more information. The pooling layer is replaced by convolution with stride 2 to preserve spatial information. We further incorporate multi-modality information for accurate brain segmentation. Quantitative evaluations and comparisons with existing methods on real MR images demonstrated the significant advantages of the proposed method in terms of both segmentation accuracy and parameter efficiency. In the future, we will explore the proposed network architecture for difficult task such as tumor segmentation.

\bibliography{main}
\bibliographystyle{splncs03}

\end{document}